%% file: publication_ICRA2027_OlivierGamache.tex
\documentclass[letterpaper, 10 pt, conference]{ieeeconf}  

\IEEEoverridecommandlockouts                              
\input{./latexGoodPractices/preamble}

\usepackage{physics}  
\AtBeginDocument{\RenewCommandCopy\qty\SI}  
\usepackage{amsmath} 
\usepackage[normalem]{ulem}
\usepackage{algorithm}
\usepackage[noend]{algpseudocode}

\makeatletter
\let\oldtheequation\theequation
\renewcommand\tagform@[1]{\maketag@@@{\ignorespaces#1\unskip\@@italiccorr}}
\renewcommand\theequation{(\oldtheequation)}
\makeatother

\title{\LARGE \bf 
    Bracketing Shows the Way: Adaptive Multi-Exposure Policy\\as Input for Robust Visual SLAM
}

\author{Olivier Gamache$^{1}$, Anthony Lavertu$^{1}$, David-Alexandre Duclos$^{1}$, Jean-Michel Fortin$^{1}$, \\ François Pomerleau$^{1}$, Philippe Giguère$^{1}$
\thanks{$^{1}$Northern Robotics Laboratory, Université Laval, Québec City, Québec, Canada
		{\texttt{\small olivier.gamache@norlab.ulaval.ca}} and \texttt{\small {philippe.giguere@ift.ulaval.ca}}}%
}

\usepackage[switch]{lineno}
\usepackage[font=small]{caption}
\usepackage{subcaption}

\acrodef{SLAM}{Simultaneous Localization And Mapping}
\acrodef{UGV}{Uncrewed Ground Vehicles}
\acrodef{UAV}{Uncrewed Aerial Vehicle}
\acrodef{IMU}{Inertial Measurement Unit}
\acrodef{CNN}{Convolutional Neural Networks}
\acrodef{TnR}{Teach-and-Repeat}
\acrodef{fps}{frames per second}
\acrodef{GNSS}{Global Navigation Satellite System}
\acrodef{BEV}{Bird's-Eye View}
\acrodef{FoV}{Field-of-View}
\acrodef{MLP}{Multi-Layer Perceptron}
\acrodef{RL}{Reinforcement Learning}
\acrodef{RMSE}{Root Mean Squared Error}
\acrodef{ATE}{Absolute Trajectory Error}
\acrodef{RPE}{Relative Pose Error}
\acrodef{DTE}{Distance Traveled Error}
\acrodef{AE}{Automatic Exposure}
\acrodef{HDR}{High Dynamic Range}
\acrodef{SLAM}{Simultaneous Localization and Mapping}
\acrodef{VSLAM}{Visual SLAM}
\acrodef{SNR}{Signal-to-Noise Ratio}
\acrodef{FPS}{Frames Per Second}
\acrodef{EV}{Exposure Value}
\acrodef{BVSLAM}{Bracketing-based Visual Simultaneous Localization and Mapping}
\acrodef{EMA}{Exponential Moving Average}

\newcommand{\methodName}{\mbox{BVSLAM}\xspace}
\newcommand{\bracketingName}{\mbox{B-Adapt}\xspace}

\newcommand{\SE}{\mathrm{\mathbf{SE}}}
\newcommand{\Low}{\texttt{Low}\xspace}
\newcommand{\Mid}{\texttt{Mid}\xspace}
\newcommand{\High}{\texttt{High}\xspace}

\begin{document}
\maketitle
\thispagestyle{empty}
\pagestyle{empty}

\begin{abstract}
\boldmath
\sisetup{text-series-to-math=true}
Rapid illumination transitions can disturb \ac{VSLAM} by saturating images before the \ac{AE} policy adapts, removing the visual correspondences needed to maintain tracking. 
Our proposed methods mitigate this disturbance at two levels: \bracketingName acquires adaptive exposure brackets to cover the scene dynamic range while promoting photometric overlap, and \methodName directly uses the multi-exposure sequence as constraints in a factor graph state estimator without relying on \ac{HDR} merging.
We evaluate both methods against state-of-the-art approaches in nominal environments and during a repeated tunnel-exit transition at an emulated vehicle speed of \qty{32}{\meter\per\second}. In the tunnel, combining our \bracketingName with \methodName reduces the state estimation error by \qty{78}{\percent} compared with \ac{AE} images. 
These results suggest that bracketing efficiently improves visual pose estimation robustness to rapid illumination disturbances while retaining low error under nominal conditions.
Our code is available at \url{https://github.com/norlab-ulaval/bracketing-shows-the-way}.
\end{abstract}


\acresetall 


\section{Introduction}
\label{sec}

\begin{figure}[t]
\centering
\includegraphics[width=0.99\linewidth]{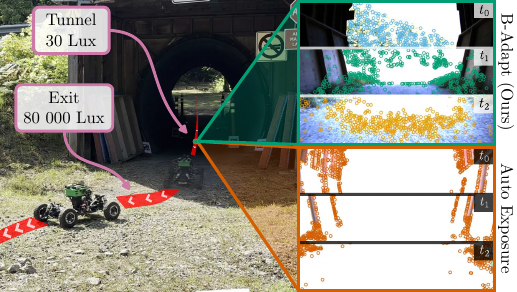}
\caption{Effect of a rapid low-to-high illumination transition on the visual information available for pose estimation. \textit{Left}: The vehicle exits a dim tunnel (\qty{30}{\lux}) toward a bright outdoor scene (\qty{80000}{\lux}). \textit{Right}: Camera images acquired by the vehicle during the tunnel-exit traversal shown on the left. The vertically cropped images show three successive instants ($t_0$--$t_2$) during the illumination transition. Colored points indicate the detected visual features. The \ac{AE} policy adapts too slowly, saturating images, whereas \bracketingName preserves observable content across complementary exposures.}
\label{fig:intro}
\vspace{-0.15in}
\end{figure}

\ac{SLAM} algorithms work well in standard environments (\emph{nominal}), but are affected by \emph{disturbances}, defined as short but meaningful perturbations that can degrade their performance~\citep{boxan2026oneyear}.
For \ac{VSLAM}, rapid illumination changes can produce such disturbances by saturating the images that constitute its primary sensing modality~\citep{zuniga2020vi}.
This limitation is particularly pronounced at a tunnel exit, where the scene dynamic range can reach \qty{120}{\decibel}~\citep{hertel2008adaptive}, compared with theoretical ranges of \qty{48}{\decibel} for \mbox{8-bit} image acquisition.
Cameras typically mitigate saturation through an \ac{AE} policy that adjusts exposure time and gain according to the observed scene.
However, standard \ac{AE} policies are reactive and therefore inherently introduce an adaptation delay~\citep{shim2014auto}. 
During rapid illumination changes, this delay can produce saturated frames, as in the high-speed tunnel exit of \cref{fig:intro}, depriving \ac{VSLAM} of visual constraints and potentially causing tracking failure.

The robustness of \ac{VSLAM} to illumination disturbances can be improved through two complementary strategies: controlling the image acquisition policy or adapting the state estimator that processes the images.
For camera pose estimation, \ac{VSLAM} approaches rely on deep learning-based feature extractors and matchers resilient to illumination changes, thereby creating better constraints for optimization~\citep{xu2025airslam}.
For image acquisition policies, existing approaches mainly modify the exposure-control heuristic~\citep{zhang2017active} or learn brightness parameters that maximize \ac{VSLAM} performance~\citep{zhang2024efficient}.
Nevertheless, these methods acquire a single exposure at each instant and remain limited by the sensor dynamic range.
To overcome this, image bracketing captures complementary exposure levels, which are conventionally merged into a \ac{HDR} image~\citep{mertens2009exposure}.
However, this reconstruction can introduce ghosting and misregistration artifacts under camera motion~\citep{yan2024dynamic}.
More recent work has demonstrated dual-exposure acquisition with a standard camera for depth estimation~\citep{choi2025dual}.
Others have used a specialized camera to acquire single-shot multi-exposure images for \ac{VSLAM}, eliminating inter-exposure motion, thereby facilitating their fusion~\citep{caillot2026dual}.
Importantly, current \ac{VSLAM} systems are not designed to directly exploit exposure brackets acquired under camera motion, leaving information preserved across exposures unavailable for pose estimation, thus increasing the risk of complete tracking loss.


In this work, we introduce both an adaptive bracketing policy (\emph{\bracketingName}) and a bracketing-based \ac{VSLAM} (\emph{\methodName}) to mitigate rapid illumination disturbances.
Our \bracketingName controls the exposure levels to cover the scene dynamic range while promoting photometric overlap between consecutive brackets.
By distributing the acquisitions across multiple exposure levels, \bracketingName preserves observable image content during rapid transitions without relying on a single exposure to converge.
Our \methodName then bypasses \ac{HDR} reconstruction through a novel factor graph formulation that directly processes the bracketed images.
We evaluate both mitigation strategies against state-of-the-art approaches in nominal conditions and during a rapid tunnel-exit disturbance, similar to that considered by \citet{Tomasi2021}, as depicted in \cref{fig:intro}.

In short, the main contributions of this paper are:
\begin{itemize}
    \item An adaptive image-bracketing policy, \emph{\bracketingName}, which acts as an \textit{acquisition-side} mitigation strategy for rapid illumination disturbances.
    \item A novel factor graph-based \ac{VSLAM} formulation, \emph{\methodName}, designed to exploit bracketed sequences as an \textit{estimation-side} mitigation strategy.
    \item An extensive analysis of both mitigation strategies in nominal and disturbance-affected regions against multi-exposure acquisition and \ac{VSLAM} approaches.
\end{itemize}

\section{Related Work}
\label{sec:related_work}

We first review acquisition-side approaches to illumination disturbances, from single-exposure control to multi-exposure bracketing under camera motion. 
We then examine how \ac{VSLAM} systems address challenging illumination scenarios, including methods that incorporate multiple exposures into pose estimation.

\subsection{Single-Exposure Acquisition for Robotics}
\label{sec:rw-hdr}

In robotics, research on \ac{AE} policies optimizes metrics related to the downstream estimation task rather than relying solely on mean image intensity.
These methods are divided into two main families based on how they select the next exposure: \emph{reactive} policies respond to the current observation, and \emph{predictive} policies anticipate the upcoming scene.
\emph{Reactive} policies for \ac{VSLAM} regulate exposure using image gradients~\citep{zhang2017active}, histogram uniformity~\citep{li2025automatic}, or a combination of gradient, entropy, and \ac{SNR}~\citep{kim2020proactive}.
Because evaluating a potential exposure setting traditionally requires physically acquiring an image, several methods bypass this issue by synthesizing virtual exposures from the current frame~\citep{kim2020proactive,han2023camera}.
\Citet{zhang2024vircap} extend this approach by directly outputting the selected synthetic image from the camera driver.
Nevertheless, \emph{reactive} policies optimize exposure from an already observed scene, introducing delays which are problematic during rapid illumination transitions.
To mitigate this delay, \emph{predictive} policies anticipate the next exposure before the scene changes, using either self-supervision~\citep{Tomasi2021} or reinforcement learning~\citep{zhang2024efficient}.
Another metric optimization policy builds a representative map of the environment and selects exposures to recover radiance values missing due to saturation in regions expected to enter the \ac{FoV}~\citep{alexandrov2017high}.
\Citet{barat2017active} also use map feedback, actively controlling the exposure time to increase the information content of a 3D \ac{HDR} environment map.
In a benchmark of \emph{reactive} and \emph{predictive} policies, \citet{gamache2025reproducible} showed that standard \ac{AE} remains a strong image-acquisition heuristic for \ac{VSLAM} tasks.
Regardless of how exposure is selected, however, each policy acquires a single image at a time and remains limited by the sensor dynamic range. 
The lack of multi-exposure approaches motivates the bracketing policy investigated in this work.

\subsection{Multi-Exposure Bracketing under Motion}
\label{rw-bracketing}

Several methods rely on brackets to form a unified representation for estimation. \Citet{zhang2024image} use darker and brighter images to update their main well-exposed image passed to \ac{VSLAM} algorithms, whereas \citet{zhang2024vircap} synthesize one image from five fixed, closely spaced exposure levels. 
Multi-exposure bracket images can instead be merged into a single frame to recover the scene dynamic range~\citep{mertens2009exposure}. 
Under motion, misalignment causes ghosting and registration artifacts, particularly where the reference frame is saturated~\citep{yan2024dynamic}.
\Citet{gupta2013fibonacci} estimate motion between sums of adjacent frame sets with equal total exposure, but report that the method may fail under extremely rapid motion or long intervals between frames.
HDRFlow~\citep{xu2024hdrflow} accommodates larger displacements using optical flow in the \ac{HDR} domain, although occluded regions are still challenging for image merging.
Wider coverage requires more or farther-separated exposures, increasing the time and viewpoint changes between brackets~\citep{barat2017active}.

Bracketing policies can also adapt to scene conditions. 
For \ac{HDR} video, \citet{gupta2013fibonacci} vary exposure spacing with scene brightness and motion, whereas \citet{gryaditskaya2015motion} select exposures to balance dynamic-range coverage against motion-related reconstruction artifacts. 
For stereo depth estimation, \citet{choi2025dual} adapt two alternating exposures to the scene dynamic range, separating them when it exceeds the camera range and bringing them closer together otherwise. 
They cap the separation to retain overlapping content for optical-flow-based alignment before stereo depth estimation.

Specialized sensors avoid sequential-acquisition viewpoint changes using spatially varying exposure~\citep{ye2022visual} or on-sensor dual-exposure readout~\citep{caillot2026dual}.
However, these sensors are not commonly available on robotic platforms, and adaptive exposure control remains an open direction~\citep{caillot2026dual}.
As opposed to these methods, our policy uses standard camera hardware to acquire bracketing image sequences and uses them as direct input to \ac{VSLAM} algorithms.

\begin{figure*}[htbp]
    \centering
    \includegraphics[width=0.99\linewidth]{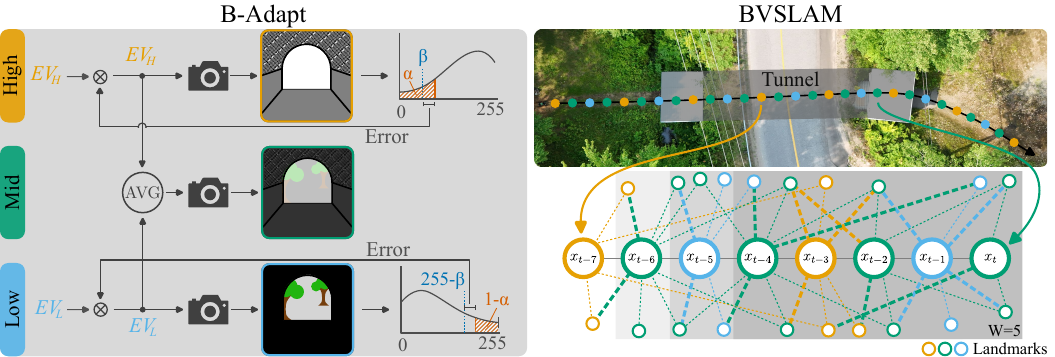}
    \caption{Overview of our proposed \bracketingName and \methodName pipeline for $n=3$, $I_{\max}=255$, and $W=5$. \textit{Left}: B-Adapt independently regulates the \Low and \High bounding brackets using one-sided intensity-tail objectives, parameterized by $\alpha$ and $\beta$, while the intermediate \Mid exposure is derived from the two bounds. \textit{Right}: BVSLAM exploits the resulting bracketing sequence by associating features of images within one exposure step and a lookback window of $W$ frames from the current one (grey window). Pose colors indicate the corresponding exposure levels, while the unfilled circles denote landmarks shared across poses through the retained cross-exposure correspondences.}
    \label{fig:metho-pipeline}
    \vspace{-0.15in}
\end{figure*}

\subsection{VSLAM under Challenging Illumination}
\label{rw-vslam}

While the preceding approaches address illumination disturbances during image acquisition, \ac{VSLAM} systems can also mitigate their effects during pose estimation. 
State-of-the-art \ac{VSLAM} algorithms, such as ORB-SLAM3~\citep{campos2021orb} and CuVSLAM~\citep{korovko2025cuvslam}, are designed and evaluated using conventional sequences of consecutive \ac{AE} images rather than bracketed image sequences. 
AirSLAM~\citep{xu2025airslam} is resilient to short and long illumination changes using learned point and line features.
\Citet{li2016hdrfusion} instead use exposure-invariant radiance measurements for tracking and accumulate observations in an \ac{HDR} map of the environment.
These estimation-side methods can improve robustness to appearance changes, but cannot restore detail missing from a single saturated observation.
A smaller body of work directly incorporates bracket image sequences into pose estimation. 
\Citet{irie2011high} combine keypoints from four sequential exposures and localize against a previously built map of the environment using odometry. 
\Citet{ye2022visual} use four simultaneous exposures from a spatially varying exposure camera for visual odometry, while \citet{caillot2026dual} use a specialized dual-exposure camera for \ac{VSLAM}.
Motivated by these limitations in both image acquisition \textit{and} pose estimation, our work combines two complementary strategies to mitigate rapid illumination disturbances: \bracketingName, an adaptive bracketing policy to cover the scene dynamic range while promoting cross-exposure photometric overlap, and \methodName for state estimation without relying on \ac{HDR} image merging.

\section{Theory}
\label{sec:methodology}

Our method circumvents bracket information merging by using multi-exposure images directly for pose estimation.
Such an approach requires both an acquisition policy that maintains informative visual features across exposure levels and a \ac{VSLAM} formulation capable of exploiting these cross-exposure observations.
In this section, we introduce our online policy \bracketingName that adapts the bracket set to the observed scene dynamic range, before describing how \methodName combines the acquired bracketed images within a factor graph formulation.

\subsection{\bracketingName: Bracketing Acquisition Policy}
\label{sec:metho-bracketing}

Prior knowledge of the scene dynamic range is unavailable during online acquisition.
As such, estimates obtained from observations may quickly become outdated as the camera moves rapidly between regions with drastically different lighting conditions.
We thus propose an adaptive acquisition policy that estimates the bounding exposure levels directly from the acquired images.
The policy controls the camera exposure time $t$ and gain $g$, whose multiplicative effects on the collected signal are represented by a single scalar state $e$ in the logarithmic \ac{EV} domain.
As illustrated in \cref{fig:metho-pipeline}, the proposed policy relies on two independent control loops focusing on opposite regions of the scene radiance, and is associated with two bounding brackets dubbed \Low and \High.
As opposed to conventional \ac{AE} policies where the optimized value is the mean image intensity, each bounding bracket is regulated using a one-sided tail objective evaluated only on images acquired using that control loop.
The objective for the \Low bracket controller is defined as
\begin{equation}
    \hat{e}_L = \underset{e_L}{\operatorname{argmin}}\left|P_{1-\alpha}\left(I(e_L)\right)-\left(I_{max}-\beta\right)\right|,
    \label{eq:target_L}
\end{equation}
where $P_q$ denotes the $q$-percentile intensity, $I(e)$ are the images acquired with the exposure value $e$, $I_{\max}$ is the maximum representable sensor intensity (e.g., 255 for \mbox{8-bit} cameras), $\alpha$ is the tolerated fraction of pixels that are excluded from the dynamic range estimation, essentially considered as outliers, and $\beta$ is an intensity margin that keeps the regulated tail away from the limits of the sensor range.
The \Low controller focuses on the bright regions of the scene, ensuring that only a fraction $\alpha$ of the pixels exceed $I_{\max}-\beta$ and approach saturation.
Conversely, the \High controller focuses on the dark regions of the scene and targets a fraction $\alpha$ of the pixels to fall below $\beta$:
\begin{equation}
    \hat{e}_H = \underset{e_H}{\operatorname{argmin}} \left|P_{\alpha}(I(e_H)) - \beta\right|.
    \label{eq:target}
\end{equation}
Each bounding bracket's next $e$ is updated using an \ac{EMA} with coefficient $\lambda$, controlling the reactiveness of the controller.

When only two brackets are used ($n=2$), the \Low and \High control loops form a complete bracket set.
For larger sets ($n\geq3$), we insert $n-2$ intermediate exposures that act as bridges between bounding brackets, thus increasing the photometric coverage.
The intermediate exposure levels evenly divide the \ac{EV} interval according to
\begin{equation}
    e_{M_k} = e_L + \frac{k}{n-1}\left( e_H - e_L \right),
    \qquad k = 1, \dots, n-2 ,
    \label{eq:mid}
\end{equation}
where $k \in \{1, 2, \ldots, n-2\}$ indexes the intermediate brackets.
The intermediate exposure levels are dynamically computed from the latest $e_L$ and $e_H$.
In low-dynamic-range scenes, the two objectives can be satisfied by similar exposure levels, causing the bounding brackets to converge toward one another, thereby increasing vulnerability to rapid illumination changes.
We therefore impose a minimum-separation constraint:
\begin{equation}
\Delta_{\min} < |e_H-e_L|.
\end{equation}
Note that the proposed policy does not impose an upper bound on the separation between \Low and \High, allowing the brackets to separate as far as needed to cover the full dynamic range of the scene.

In order to maximize the photometric overlap between consecutive frames, the acquisition sequence should transition only between adjacent exposure levels.
Following \citet{zhang2024image}, we therefore acquire the brackets using a specific sequence ${\Low, \Mid_1, \dots, \Mid_{k}, \High, \Mid_{k}, \dots, \Mid_1}$, as illustrated on the bottom right side of \cref{fig:metho-pipeline} showing an example for $n=3$.
This ordering prevents \Low and \High, the two least photometrically similar brackets, from being acquired consecutively when $n\geq3$, which would result in a large, detrimental jump in exposure.

The resulting acquisition policy preemptively covers multiple exposure levels and is therefore designed to accommodate rapid illumination changes. In contrast, conventional \ac{AE} assumes sufficiently gradual changes and prioritizes the visual appearance of each image. By instead prioritizing the preservation of usable visual information in difficult lighting conditions, the proposed bracketing policy is designed to facilitate localization and mapping tasks.

\subsection{\methodName Implementation}
\label{sec:metho-slam}


Current \ac{VSLAM} algorithms are not designed to natively process multi-exposure bracketed sequences. 
We thus propose \ac{BVSLAM}, which exploits a known bracketing cycle to construct visual constraints using the illumination-resilient SuperPoint detector~\citep{detone2018superpoint} and LightGlue matcher~\citep{lindenberger2023lightglue}. 
By linking landmarks across exposure levels, the factor graph jointly constrains the camera poses directly from the bracketed sequence, without intermediate image fusion.
Each image is matched with every other image located within one exposure step of itself and contained within a lookback window of $W$ frames, including the current pose, as illustrated on the bottom right side of \cref{fig:metho-pipeline}.
Restricting the associations in this manner avoids matching images with large exposure differences, whose limited photometric overlap could produce unreliable correspondences.
We select $W$ as a multiple of the bracketing-cycle period, ensuring that the temporal association window represents the complete acquisition cycle.


The retained stereo and temporal correspondences form the visual constraints of the state-estimation problem. 
We formulate the problem as a factor graph over the camera poses $\mathbf{X}i\in\SE(3)$ and landmark positions $\mathbf{p}{\ell}$ obtained from the image matches, using GTSAM~\citep{gtsam}. 
Typical \ac{VSLAM} systems employ a kinematic motion model to propagate pose estimates when visual constraints become weak. 
We instead deliberately apply a zero-motion prior between consecutive poses, preventing the estimator from propagating the camera trajectory through regions with insufficient visual information and thereby isolating the effect of the acquisition policies on the visual constraints available for pose estimation.
The resulting nonlinear least-squares problem is first solved incrementally over a sliding window and subsequently refined over the complete graph using global bundle adjustment.

The current \methodName implementation is intended for offline processing and does not operate in real time.
The primary objective of the implementation is to evaluate the influence of image bracketing on the visual constraints used for pose estimation rather than to achieve real-time performance. 

\section{Methodology}
\label{sec:metho}

We evaluate the proposed acquisition and state-estimation methods in both nominal conditions and during a rapid illumination increase, termed the disturbance condition. 
This section first presents the acquisition platform and implementation parameters, then describes the recorded environments, and finally details the compared methods, reference trajectories, and evaluation metrics.

\begin{figure*}
    \centering
    \includegraphics[width=0.99\linewidth]{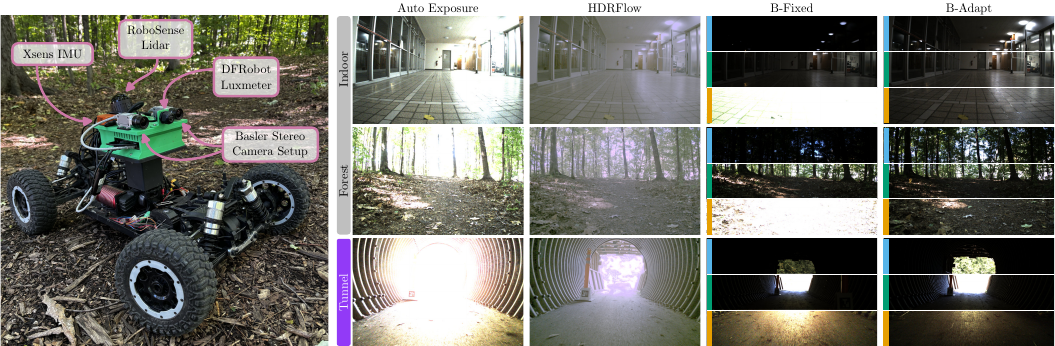}
    \caption{\textit{Left}: Our experimental platform.
    The acquisition head combines a synchronized stereo pair with sensors used to compute reference trajectories and characterize scene illumination.
    \textit{Right}: Examples of frames taken across our nominal (grey) and disturbance (purple) trajectories, with the evaluated approaches.
    \bracketingName and B-Fixed images are stacked and cropped horizontally to showcase the three image exposures.}
    \label{fig:experiments-setup}
    \vspace{-0.15in}
\end{figure*}

\subsection{Experimental Setup}
\label{sec:experiments-setup}

We collected the experimental data using the acquisition head shown in green in \cref{fig:experiments-setup}. The head carries a custom hardware-synchronized stereo pair of global-shutter Basler a2A1920-51gcPRO cameras. A Robosense Airy \num{128}-beam lidar and an Xsens MTi-100 \ac{IMU} provide the measurements used to compute the reference trajectories, while a DFRobot SEN0644 luxmeter records the illumination across the environments. 
Our custom \bracketingName policy drives the stereo pair using the parameters reported in \cref{tab:acquisition-parameters}. 
The exposure-time and gain bounds were selected to balance motion-blur reduction against sufficient image brightness.
The acquisition head was mounted on a human-operated 1/5-scale remote-controlled car. 
We physically limited the throttle command to maintain an approximately constant speed along each trajectory. Since the vehicle cannot reach full-scale road-vehicle speeds, we follow an evaluation approach similar to \citet{bednavr2022deployment} and reduce the camera \ac{FPS} to reproduce the larger inter-frame displacements associated with high-speed motion.

\begin{table}[htpb]
    \centering
    \small
    \caption{\bracketingName parameters used in our experiments.}
    \begin{tabular*}{\linewidth}{@{\extracolsep{\fill}} l l l }
        \toprule
        \textbf{Parameter} & \textbf{Symbol} & \textbf{Value} \\
        \midrule
        Number of brackets & $n$ & $3$ \\
        Window size & $W$ & $5$ \\
        Tail fraction & $\alpha$ & \qty{1.0}{\percent} \\
        Intensity margin & $\beta$ & $5$ \\
        Smoothing coefficient & $\lambda$ & $0.8$ \\
        Minimum separation & $\Delta_{\min}$ & $1~\text{stop}$ \\
        Exposure-time range & $t$ & $[\qty{19}{\micro\second},\qty{5}{\milli\second}]$ \\
        Gain range & $g$ & $[\qty{0}{\decibel},\qty{48}{\decibel}]$ \\
        \bottomrule
    \end{tabular*}\label{tab:acquisition-parameters}
\end{table}

\subsection{Experiment Environments}
\label{sec:metho-env}
We evaluate \bracketingName and \methodName under two operating regimes: nominal conditions representative of standard \ac{VSLAM} operation, and a rapid \ac{HDR} transition used as an extreme illumination disturbance. 
The nominal regime comprises the \emph{Indoor} and \emph{Forest} trajectories shown in the top right of \cref{fig:experiments-setup}, for a total of \SI{650}{\meter} of recorded data. 
These sequences were acquired at an approximate vehicle speed of \qty{2.3}{\meter\per\second} with the camera operating at $f=\qty{20}{\hertz}$. 
For the disturbance regime, we use the under-road \emph{Tunnel} shown at the bottom of \cref{fig:experiments-setup}, following an experimental setting similar to \citet{Tomasi2021}, and focus on the tunnel exit, where the rapid low-to-high illumination transition combined with a high traversal speed consistently challenges the exposure policies. 
By contrast, the high-to-low transition at the tunnel entry occurs more gradually and does not disrupt the evaluated acquisition policies. 
We therefore do not consider this transition a disturbance.
The resulting trajectory spans \qty{55}{\meter}, including approximately \qty{21}{\meter} inside the tunnel, and was traversed at an average speed of \qty{4.3}{\meter\per\second}. 
To further increase the difficulty of the transition, we reduce the camera frame rate to $f=\qty{4}{\hertz}$, yielding an inter-frame displacement equivalent to a vehicle traveling at approximately \qty{32}{\meter\per\second} with a \qty{30}{\hertz} camera.

\subsection{Evaluation Modalities}
\label{sec:metho-eval}

The evaluation accounts for the two strategies of mitigation investigated in this work: image acquisition and state estimation. 
At the acquisition level, we compare \bracketingName with three alternatives as depicted in \cref{fig:experiments-setup}. 
\ac{AE} provides the single-exposure baseline~\citep{gamache2025reproducible}, whereas HDRFlow~\citep{xu2024hdrflow} merges the brackets acquired by \bracketingName, eliminating saturated regions but introducing distortions and artifacts in the reconstructed scene due to large motion between frames.
B-Fixed serves as an ablation in which the exposure time and gain of the \Low and \High brackets are fixed to cover the full dynamic range of the tunnel environment. 
Other interesting policies were also considered for benchmarking against our approach, but their implementations are not publicly available~\citep{xu2025adaptiveae, Tomasi2021}.
Because the platform has only one stereo pair, we independently recorded each acquisition policy five times over the same trajectory for the disturbance scenario to mitigate inter-trajectory variations, whereas each nominal trajectory was recorded once per policy.

At the state-estimation level, we process the four acquisition outputs with \methodName and compare their performance with ORB-SLAM3~\citep{campos2021orb}, CuVSLAM~\citep{korovko2025cuvslam} and AirSLAM~\citep{xu2025airslam}. 
The parameters of each \ac{VSLAM} algorithm were tuned for the acquisition setup and environments. 
The resulting trajectories are evaluated against reference trajectories computed using the lidar-inertial odometry method of \citet{kubelka2022gravity}, based on \texttt{libpointmatcher}.\footnote{\url{https://github.com/norlab-ulaval/libpointmatcher}} 
Relying exclusively on lidar and inertial measurements makes the reference solution independent of the illumination transitions affecting the visual estimators.

Based on our camera frame rate and the vehicle's real speed, trajectory accuracy is quantified using the \ac{RPE} over \qty{5}{\meter} windows. 
For the disturbance regime, the windows from all five repetitions are pooled, and their \ac{RMSE} is reported. 
The \ac{RMSE} emphasizes the large errors associated with severe trajectory deviations or tracking loss more strongly than median- or mean-based summaries, making it particularly relevant for evaluating robustness to localized disturbances. 
All trajectory metrics are computed using the \texttt{evo} package.\footnote{\url{https://github.com/MichaelGrupp/evo}}


\section{Results}
\label{sec:results}

We first assess how the proposed acquisition and state-estimation policies affect \ac{VSLAM} performance in nominal environments and during the rapid illumination transition disturbance. 
We then examine image saturation and feature correspondences to explain the behavior observed during the disturbance.

\subsection{VSLAM Analysis on Nominal and Disturbance Datasets}
\label{sec:results-slam}

\begin{figure}[t]
\centering
\includegraphics[width=0.99\linewidth]{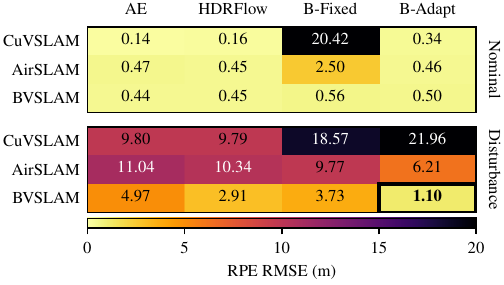}
\caption{\ac{RPE} \ac{RMSE} over \qty{5}{\meter} windows in the nominal environments (\textit{top}) and during the tunnel-exit disturbance (\textit{bottom}), evaluated across state estimators (\textit{rows}) and acquisition policies (\textit{columns}). Disturbance results are pooled across five repeated trajectories. The outlined cell identifies our proposed combination of \bracketingName and \methodName.}
\label{fig:results-slam}
\vspace{-0.15in}
\end{figure}

We compare each acquisition policy and state estimator in nominal environments and during the Tunnel disturbance using the \ac{RPE} \ac{RMSE}, as shown in the upper and lower matrices of \cref{fig:results-slam}, respectively.
We omit ORB-SLAM3 from the figure for clarity because it performed adequately in nominal environments but lost tracking in every Tunnel run, consistent with the difficulty of comparable illumination transitions reported by \citet{Tomasi2021}.

Under nominal conditions, all three estimators perform similarly using the \ac{AE} policy as baseline, including \methodName.
Their average \ac{RPE} \ac{RMSE} corresponds to \qty{7}{\percent} of the \qty{5}{\meter} evaluation interval, which is similar to the results reported in \citet{boxan2026oneyear} for \ac{VSLAM} algorithms.
Fixed bracketing, however, substantially increases the error of CuVSLAM and AirSLAM.
The wide photometric separation between its bracket exposure levels leaves little photometric overlap between successive brackets, making their association difficult for estimators designed for conventional image sequences. 
\methodName can associate observations across exposure levels and retains an error of the same order as with \ac{AE}. 
With \bracketingName, the brackets move closer together in scenes with narrower dynamic range, and all three estimators retain low nominal errors. 
Thus, adapting the bracket spacing preserves the applicability of multi-exposure acquisition under standard conditions.

For the disturbance scenario, CuVSLAM and AirSLAM reach errors of \qty{9.80}{\meter} and \qty{11.04}{\meter}, around \qty{70} and \qty{23} times their respective nominal errors when relying on the \ac{AE} policy. 
These increases are associated with tracking loss at the tunnel exit. 
\methodName obtains a lower error than other estimators with the same \ac{AE} policy inputs. Our method benefits from its factor graph formulation that uses global bundle adjustment and the absence of simplification required for real-time operation present in other state-of-the-art methods.
Among the acquisition outputs, \bracketingName yields the lowest error on the tunnel trajectory for \methodName at \qty{1.10}{\meter}, compared with \qty{4.97}{\meter} using \ac{AE}, \qty{3.73}{\meter} using B-Fixed, and \qty{2.91}{\meter} using HDRFlow. 
The \bracketingName result represents a \qty{78}{\percent} reduction relative to \ac{AE} and remains comparable to the performance obtained with the same combination in nominal environments. 
Compared with B-Fixed and HDRFlow, adaptive bracketing achieves lower error without requiring prior scene knowledge or computationally expensive HDR merging.
Similar to the conclusions of \citet{ye2022visual}, we note better performance for direct use of the images acquired using the \bracketingName policy rather than merging with HDRFlow.
Interestingly, although AirSLAM is not designed for direct use of multi-exposure image sequences, it achieves its best result for the disturbance case with \bracketingName, at \qty{6.21}{\meter} error, and avoids the tracking loss observed with the \ac{AE} policy as input. 
Overall, \bracketingName improves the robustness to disturbance for AirSLAM and \methodName, with the largest reduction observed for \methodName.

Overall, neither B-Fixed nor HDRFlow improves upon \bracketingName when processed by \methodName, while \ac{AE} remains a competitive and widely applicable single-exposure baseline, consistent with the findings of \citet{gamache2025reproducible}. We therefore focus the remainder of the analysis on \ac{AE} and \bracketingName to isolate how adaptive bracketing preserves visual information relative to conventional image acquisition.

\subsection{Exposure Policies in Rapid Illumination Transition}
\label{sec:results-behavior}

\begin{figure*}[t]
\centering
\includegraphics[width=0.99\linewidth]{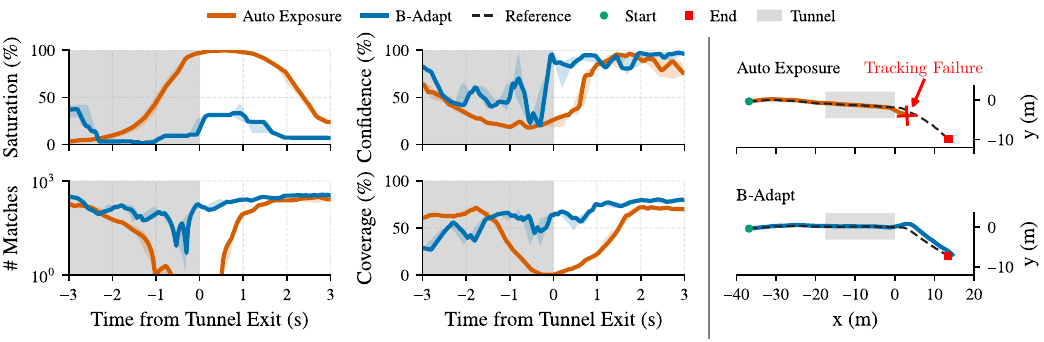}
\caption{
Comparison of \ac{AE} and \bracketingName around the tunnel exit. \emph{Left}: The plots report image saturation, median LightGlue confidence, number of SuperPoint--LightGlue matches, and feature coverage over a rolling window of five images. Feature coverage measures the proportion of the current frame covered by features matched and reprojected from previous images in the window. Only matches with confidence above $0.9$ are retained. Solid lines and shaded regions represent the median and IQR across five repetitions. \emph{Right}: \methodName trajectories obtained using \ac{AE} and \bracketingName, together with the reference trajectory.
}
\label{fig:results-expo-policies}
\vspace{-0.15in}
\end{figure*}

To examine the visual constraints available to \methodName at the tunnel exit, \cref{fig:results-expo-policies} compares \ac{AE} and \bracketingName policies along the trajectory. 
In addition, the right panel shows the bird's-eye view of one of the tunnel trajectories computed from \ac{AE} and \bracketingName acquisition policies using \methodName.
At the low-to-high illumination transition, the \ac{AE} images become almost entirely overexposed. 
The saturation ratio approaches \qty{100}{\percent}, while high-confidence matches fall to nearly zero for approximately one second, leaving few usable visual constraints.
As described in the \methodName implementation, our study isolates the effect of the visual disturbance on camera-based \ac{SLAM} by excluding kinematic motion models and nonvisual sensing modalities.
Therefore, the lack of visual information is the main reason why \ac{VSLAM} algorithms often lose tracking during rapid illumination transitions, having nothing to estimate the displacement of the camera in the environment.
\bracketingName also exhibits a brief adaptation delay, visible in the saturation trace, but its pre-existing exposure span allows the \Low bracket to retain observable content through the exit. 
Reducing $\alpha$ and $\beta$ or increasing $\lambda$ could shorten this delay, although it may compromise the stability of the exposure policy. 
Consequently, with \bracketingName, informative correspondences are found even during the rapid transition.

The trajectories depicted in \cref{fig:results-expo-policies} qualitatively illustrate the effect of the disturbance on pose estimation. 
The \ac{AE}-based estimate accurately follows the reference until the disturbance, where it completely loses track near the exit of the tunnel. 
Meanwhile, the \bracketingName-based estimate shows only a small deviation at the disturbance site, consistent with the visual information retained by its \Low bracket and with the lower observed disturbance error in \cref{fig:results-slam}.


Combining \bracketingName with \methodName retains comparable performance in nominal conditions, while preserving sufficient visual information to substantially outperform the evaluated alternatives during the tunnel-exit disturbance at an emulated speed of \qty{32}{\meter\per\second}.

\section{Discussion}
\label{sec:discussion}

The results demonstrate that combining \bracketingName with \methodName improves robustness to rapid illumination transitions by exploiting the complete exposure cycle. 
We next examine how this benefit evolves with emulated vehicle speed before discussing the limitations of our proposed approach.

\subsection{Effect of Emulated Speed on VSLAM Performance}
\label{sec:discussion-fps}

\begin{figure}[htpb]
    \centering
    \includegraphics[width=0.99\linewidth]{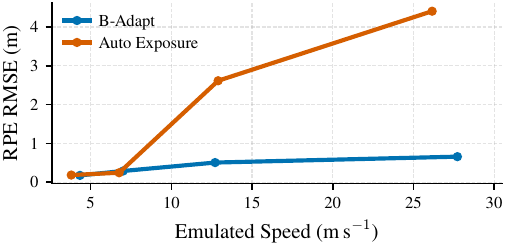}
    \caption{Effect of emulated vehicle speed on \ac{RPE} \ac{RMSE}. Emulated speed is computed relative to a \qty{30}{\hertz} camera and increased by lowering the frame rate. Results are pooled across five repeated trajectories for each acquisition policy and frame-rate setting.}
    \label{fig:results-fps}
    \vspace{-0.15in}
\end{figure}

To assess performance beyond the rapid emulated speed used in the disturbance sequence, we varied the camera frame rate and compared \ac{AE} with \bracketingName, both processed by \methodName. 
As shown in \cref{fig:results-fps}, the policies produce similar trajectory errors at lower emulated speeds. 
As emulated speed increases, the \ac{AE} policy has fewer image observations over which to converge during the illumination transition, causing its error to rise sharply, whereas \bracketingName remains comparatively stable.
This divergence is consistent with the loss of feature correspondences observed for \ac{AE} at the Tunnel exit in \cref{fig:results-expo-policies}.
The \ac{AE} policy adaptation delay becomes more consequential as the camera travels farther between exposure updates.
Across the evaluated frame-rate settings, \bracketingName performs at least as well as \ac{AE}, making it a robust solution for \ac{HDR} transitions.

\subsection{Limitations}
\label{sec:discussion-limitations}
By continuously adapting distinct exposure levels, our \bracketingName policy retains usable visual information during the evaluated transition. 
However, its controllers still rely on observed images and cannot respond immediately to changes in the scene. 
Additionally, a region with an extreme intensity may initially occupy too little of the image to shift the bounding bracket controller, further contributing to the delay. 
The minimum separation between brackets maintains distinct exposure levels, anticipating sharp transitions and mitigating the resulting adaptation delay. 
Tuning $\alpha$, $\beta$, and $\lambda$ may improve responsiveness in a given environment, although faster adaptation must be balanced against policy stability.
Finally, our \methodName algorithm is limited by its computation speed, as it requires offline processing because of its overall runtime and global bundle adjustment. 
Its relatively simple factor graph formulation may allow a real-time implementation, as suggested by previous works~\citep{campos2021orb, xu2025airslam}.

\section{Conclusion}
\label{sec:conclusion}

In this work, we investigated rapid illumination transitions as disturbances to \ac{VSLAM} and proposed two complementary methods for preserving visual constraints. \bracketingName controls the exposure brackets to cover the observed scene dynamic range while promoting photometric overlap, whereas \methodName directly incorporates the resulting multi-exposure images as constraints in a factor graph state estimator. 
The combination of our \bracketingName and \methodName performs on par with or better than the evaluated state-of-the-art approaches across nominal and disturbance conditions, achieving the lowest error during the tunnel exit.
Its advantage becomes increasingly pronounced at higher emulated speeds, as the adaptation delay of the \ac{AE} policy becomes more consequential. Future work will focus on improving the responsiveness of \bracketingName while maintaining controller stability and reducing the computational cost of \methodName toward real-time operation.

\section*{Acknowledgment}


This work was supported by the Canada Research Chair on Field Robotics (CRC-2023-00396), the NSERC project CRYOTIC (ALLRP 585289-23), and the FRQNT grant 334388.

Generative AI (Claude, Anthropic) was used to assist in generating the plotting code for the data-driven result figures in \cref{sec:results}; the underlying data, analysis, and figures were produced and verified by the authors.

\printbibliography

\end{document}

%% file: latexGoodPractices/preamble.tex
\usepackage[l2tabu,orthodox]{nag}

\usepackage[
    backend=bibtex8,
    style=ieee,
    sorting=none,
    natbib=true,
    doi=false,
    isbn=false,
    url=false,
    eprint=false,
    maxcitenames=1,
    mincitenames=1
]{biblatex}

\usepackage[pdftex,colorlinks]{hyperref}

\usepackage[printonlyused]{acronym}

\usepackage{siunitx}
\usepackage[all]{nowidow}

\usepackage[dvipsnames]{xcolor}

\usepackage{lipsum}

\usepackage{xspace} 

\usepackage[pdftex]{graphicx}

\usepackage{epstopdf}

\usepackage{import}
\usepackage{svg}

\graphicspath{{./latexGoodPractices/}}

\usepackage{booktabs}

\usepackage{tabularx}
\usepackage{multirow, multicol}

\usepackage{amssymb,amsfonts,amsmath,amscd}

\usepackage{bm}

\newcommand{\bbm}{\begin{bmatrix}}
\newcommand{\ebm}{\end{bmatrix}}

\usepackage[capitalize,french,english,noabbrev,nameinlink]{cleveref}

%% file: references.bib
@article{boxan2026oneyear,
  title = {{One year in a forest: Analyzing the challenges of autonomous navigation in subarctic environments}},
      author={Matěj Boxan and Nicolas Lauzon and Veronica Vannini and Mathis Turgeon-Roy and François Pomerleau},
      year={2026},
      journal={arXiv preprint arXiv:2608.27628}
}

@inproceedings{hertel2008adaptive,
  title = {{An adaptive multiple-reset CMOS wide dynamic range imager for automotive vision applications}},
  author={Hertel, Dirk and Betts, Andrew and Hicks, Rich and ten Brinke, Machiel},
  booktitle = {Intelligent Vehicles Symposium},
  pages={614--619},
  year={2008},
  publisher = {IEEE},
}

@inproceedings{shim2014auto,
  title = {{Auto-adjusting camera exposure for outdoor robotics using gradient information}},
  author={Shim, Inwook and Lee, Joon-Young and Kweon, In So},
  booktitle = {IROS},
  pages={1011--1017},
  year={2014},
  publisher = {IEEE},
}

@article{kim2020proactive,
  title = {{Proactive Camera Attribute Control Using Bayesian Optimization for Illumination-Resilient Visual Navigation}},
  author={Kim, Joowan and Cho, Younggun and Kim, Ayoung},
  journal = {IEEE Transactions on Robotics},
  volume={36},
  number={4},
  pages={1256--1271},
  year={2020},
  publisher={IEEE}
}

@article{zuniga2020vi,
  title = {{The UMA-VI dataset: Visual--inertial odometry in low-textured and dynamic illumination environments}},
  author={Zu{\~n}iga-No{\"e}l, David and Jaenal, Alberto and Gomez-Ojeda, Ruben and Gonzalez-Jimenez, Javier},
  journal={IJRR},
  volume={39},
  number={9},
  pages={1052--1060},
  year={2020},
  publisher={Sage Publications Sage UK: London, England}
}

@inproceedings{mertens2009exposure,
  title = {{Exposure Fusion: A Simple and Practical Alternative to High Dynamic Range Photography}},
  author={Mertens, Tom and Kautz, Jan and Van Reeth, Frank},
  booktitle = {Computer Graphics Forum},
  volume={28},
  number={1},
  pages={161--171},
  year={2009},
}

@article{yan2024dynamic,
  title = {{From Dynamic to Static: Stepwisely Generate HDR Image for Ghost Removal}},
  author={Yan, Qingsen and Yang, Kangzhen and Hu, Tao and Chen, Genggeng and Dai, Kexin and Wu, Peng and Ren, Wenqi and Zhang, Yanning},
  journal = {IEEE Transactions on Circuits and Systems for Video Technology},
  volume={35},
  number={2},
  pages={1409--1421},
  year={2024},
  publisher={IEEE}
}

@inproceedings{gryaditskaya2015motion,
  title = {{Motion Aware Exposure Bracketing for HDR Video}},
  author={Gryaditskaya, Yulia and Pouli, Tania and Reinhard, Erik and Myszkowski, Karol and Seidel, Hans-Peter},
  booktitle = {Computer Graphics Forum},
  volume={34},
  number={4},
  pages={119--130},
  year={2015},
}

@inproceedings{bednavr2022deployment,
  title = {{Deployment of Reliable Visual Inertial Odometry Approaches for Unmanned Aerial Vehicles in Real-world Environment}},
  author={Bedn{\'a}{\v{r}}, Jan and Petrl{\'\i}k, Mat{\v{e}}j and Vivaldini, Kelen Cristiane Teixeira and Saska, Martin},
  booktitle = {ICUAS},
  pages={167--176},
  year={2022},
  publisher = {IEEE},
}

@article{zhang2024efficient,
  title = {{Efficient Camera Exposure Control for Visual Odometry via Deep Reinforcement Learning}},
  author={Zhang, Shuyang and He, Jinhao and Zhu, Yilong and Wu, Jin and Yuan, Jie},
  journal = {RA-L},
  year={2024},
  publisher={IEEE}
}

@article{li2025automatic,
  title = {{Automatic exposure strategy network for robust visual odometry in environments with high dynamic range}},
  author={Li, Jieyu and Zhen, Ruiwen and Stevenson, Robert},
  journal={Machine Vision and Applications},
  volume={36},
  number={1},
  pages={14},
  year={2025},
  publisher={Springer}
}

@article{Tomasi2021,
   author = {Justin Tomasi and Brandon Wagstaff and Steven L. Waslander and Jonathan Kelly},
   doi = {10.1109/LRA.2021.3058909},
   issn = {2377-3766},
  journal = {RA-L},
   pages = {2028-2035},
   publisher = {IEEE},
  title = {{Learned Camera Gain and Exposure Control for Improved Visual Feature Detection and Matching}},
   number={2},
   volume = {6},
   url = {https://ieeexplore.ieee.org/document/9353970/},
   year = {2021},
}

@inproceedings{xu2025adaptiveae,
  title = {{AdaptiveAE: An Adaptive Exposure Strategy for HDR Capturing in Dynamic Scenes}},
  author={Xu, Tianyi and Zhang, Fan and Shi, Boxin and Xue, Tianfan and Wang, Yujin},
  booktitle = {ICCV},
  pages={25176--25185},
  year={2025},
  publisher = {IEEE},
}

@article{han2023camera,
  title = {{Camera Attributes Control for Visual Odometry With Motion Blur Awareness}},
  author={Han, Bin and Lin, Yicheng and Dong, Yan and Wang, Hao and Zhang, Tao and Liang, Chengyuan},
  journal = {IEEE/ASME Transactions on Mechatronics},
  volume={28},
  number={4},
  pages={2225--2235},
  year={2023},
  publisher={IEEE}
}

@inproceedings{li2016hdrfusion,
  title = {{HDRFusion: HDR SLAM Using a Low-Cost Auto-Exposure RGB-D Sensor}},
  author={Li, Shuda and Handa, Ankur and Zhang, Yang and Calway, Andrew},
  booktitle = {International Conference on 3D Vision},
  pages={314--322},
  year={2016},
  publisher = {IEEE},
}

@inproceedings{alexandrov2017high,
  title = {{High Dynamic Range SLAM with Map-Aware Exposure Time Control}},
  author={Alexandrov, Sergey V and Prankl, Johann and Zillich, Michael and Vincze, Markus},
  booktitle = {International Conference on 3D Vision},
  pages={48--56},
  year={2017},
  publisher = {IEEE},
}

@inproceedings{irie2011high,
  title = {{A High Dynamic Range vision approach to outdoor localization}},
  author={Irie, Kiyoshi and Yoshida, Tomoaki and Tomono, Masahiro},
  booktitle = {ICRA},
  pages={5179--5184},
  year={2011},
  publisher = {IEEE},
}

@inproceedings{gupta2013fibonacci,
  doi = {10.1109/iccv.2013.186},
  url = {https://doi.org/10.1109/iccv.2013.186},
  year = {2013},
  author = {Mohit Gupta and Daisuke Iso and Shree K. Nayar},
  title = {{Fibonacci Exposure Bracketing for High Dynamic Range Imaging}},
  booktitle = {ICCV},
  publisher = {IEEE},
}

@article{caillot2026dual,
  title = {{Dual-Exposure for Improving Omnidirectional Visual Simultaneous Localization and Mapping}},
  author={Caillot, Antoine and Goichon, Eva and Sagawa, Ryusuke and Caron, Guillaume},
  journal = {IEEE Robotics and Automation Practice},
  year={2026},
  publisher={IEEE}
}

@inproceedings{ye2022visual,
  title = {{Visual Odometry in HDR Environments by Using Spatially Varying Exposure Camera}},
  author={Ye, Keyang and Gao, Liuzheng and Guan, Banglei},
  booktitle = {IROS},
  pages={5995--6000},
  year={2022},
  publisher = {IEEE},
}

@inproceedings{zhang2024image,
  title = {{An Image Acquisition Scheme for Visual Odometry based on Image Bracketing and Online Attribute Control}},
  author={Zhang, Shuyang and He, Jinhao and Xue, Bohuan and Wu, Jin and Yin, Pengyu and Jiao, Jianhao and Liu, Ming},
  booktitle = {ICRA},
  pages={381--387},
  year={2024},
  publisher = {IEEE},
}

@article{zhang2024vircap,
  title = {{VirCap: Virtual Camera Exposure Control Based on Image Photometric Synthesis for Visual SLAM Application}},
  author={Zhang, Shuyang and He, Jinhao and Yang, Bowen and Zhu, Yilong and Wu, Jin and Jiao, Jianhao and Yuan, Jie},
  journal = {IEEE/ASME Transactions on Mechatronics},
  volume={30},
  number={4},
  pages={2654--2663},
  year={2024},
  publisher={IEEE}
}

@inproceedings{choi2025dual,
  title = {{Dual Exposure Stereo for Extended Dynamic Range 3D Imaging}},
  author={Choi, Juhyung and Kim, Jinnyeong and Choi, Seokjun and Lee, Jinwoo and Brucker, Samuel and Bijelic, Mario and Heide, Felix and Baek, Seung-Hwan},
  booktitle = {CVPR},
  pages={6283--6293},
  year={2025},
  publisher = {IEEE},
}

@inproceedings{barat2017active,
  title = {{Active high dynamic range mapping for dense visual SLAM}},
  author={Barat, Christian and Comport, Andrew I},
  booktitle = {IROS},
  pages={6514--6519},
  year={2017},
  publisher = {IEEE},
}

@inproceedings{zhang2017active,
  title = {{Active exposure control for robust visual odometry in HDR environments}},
  author={Zhang, Zichao and Forster, Christian and Scaramuzza, Davide},
  booktitle = {ICRA},
  pages={3894--3901},
  year={2017},
  publisher = {IEEE},
}

@software{gtsam,
  author       = {Frank Dellaert and GTSAM Contributors},
  title        = {borglab/gtsam},
  year         = 2022,
  publisher    = {Georgia Tech Borg Lab},
  version      = {4.2a8},
  doi          = {10.5281/zenodo.5794541},
  url          = {https://github.com/borglab/gtsam}
}

@article{xu2025airslam,
  title = {{AirSLAM: An Efficient and Illumination-Robust Point-Line Visual SLAM System}},
  author={Xu, Kuan and Hao, Yuefan and Yuan, Shenghai and Wang, Chen and Xie, Lihua},
  journal = {IEEE Transactions on Robotics},
  volume={41},
  pages={1673--1692},
  year={2025},
  publisher={IEEE}
}

@article{campos2021orb,
  title = {{ORB-SLAM3: An Accurate Open-Source Library for Visual, Visual--Inertial, and Multimap SLAM}},
  author={Campos, Carlos and Elvira, Richard and Rodr{\'\i}guez, Juan J G{\'o}mez and Montiel, Jos{\'e} MM and Tard{\'o}s, Juan D},
  journal = {IEEE Transactions on Robotics},
  volume={37},
  number={6},
  pages={1874--1890},
  year={2021},
  publisher={IEEE}
}

@article{korovko2025cuvslam,
  title = {{cuVSLAM: CUDA accelerated visual odometry and mapping}},
  author={Korovko, Alexander and Slepichev, Dmitry and Efitorov, Alexander and Dzhumamuratova, Aigul and Kuznetsov, Viktor and Rabeti, Hesam and Biswas, Joydeep and Pouya, Soha},
  journal={arXiv preprint arXiv:2506.04359},
  year={2025}
}

@article{gamache2025reproducible,
  title = {{Reproducible Evaluation of Camera Auto-Exposure Methods in the Field: Platform, Benchmark, and Lessons Learned}},
  author={Gamache, Olivier and Fortin, Jean-Michel and Boxan, Mat{\v{e}}j and Pomerleau, Fran{\c{c}}ois and Gigu{\`e}re, Philippe},
  journal = {T-FR},
  volume={2},
  pages={270--287},
  year={2025},
  publisher={IEEE}
}

@inproceedings{detone2018superpoint,
  title = {{SuperPoint: Self-Supervised Interest Point Detection and Description}},
  author={DeTone, Daniel and Malisiewicz, Tomasz and Rabinovich, Andrew},
  booktitle = {CVPR Workshops},
  pages={337--349},
  year={2018},
  publisher = {IEEE},
}

@inproceedings{lindenberger2023lightglue,
  title = {{LightGlue: Local Feature Matching at Light Speed}},
  author={Lindenberger, Philipp and Sarlin, Paul-Edouard and Pollefeys, Marc},
  booktitle = {ICCV},
  pages={17581--17592},
  year={2023},
  publisher = {IEEE},
}

@inproceedings{xu2024hdrflow,
  title = {{HDRFlow: Real-Time HDR Video Reconstruction with Large Motions}},
  author={Xu, Gangwei and Wang, Yujin and Gu, Jinwei and Xue, Tianfan and Yang, Xin},
  booktitle = {CVPR},
  pages={24851--24860},
  year={2024},
  publisher = {IEEE},
}

@inproceedings{kubelka2022gravity,
  title = {{Gravity-constrained point cloud registration}},
  author={Kubelka, Vladim{\'\i}r and Vaidis, Maxime and Pomerleau, Fran{\c{c}}ois},
  booktitle = {IROS},
  pages={4873--4879},
  year={2022},
  publisher = {IEEE},
}
